\documentclass[letterpaper]{article}

\usepackage{aaai2027}

\usepackage[hyphens]{url}
\usepackage{graphicx}
\usepackage{natbib}
\usepackage{caption}

\usepackage{amsmath}
\usepackage{amsfonts}
\usepackage{amssymb}

\usepackage{booktabs}
\usepackage{array}
\newcolumntype{L}[1]{>{\raggedright\arraybackslash}p{#1}}

\usepackage{algorithm}
\usepackage{algorithmic}

\usepackage{listings}
\nocopyright

\title{
When Specifications Conflict: A Symmetry-Based Framework for Measuring LLM Preferences
}

\author{
Tairan Wang\textsuperscript{*},
Liang Zhou,
Zikang Zhan\textsuperscript{*},
Pingchuan Yan
}

\affiliations{
Department of Computer Science\\
University College London\\
London, United Kingdom\\
\{tairan.wang.22, liang.zhou.22, zikang.zhan.22, maestro.yan.22\}@ucl.ac.uk\\
\textsuperscript{*}Corresponding authors.
}

\begin{document}

\maketitle

\begin{abstract}
Large language models (LLMs) are increasingly required to integrate multiple
sources of information that may be inconsistent or even conflicting. However,
there is still a lack of controllable and attributable methods for
systematically analyzing how models resolve conflicts between competing
specifications. We propose a controlled experimental framework for studying
model preferences under conflicting specifications. By constructing
specifications with explicit conflicts, the framework enables model choices
between competing specifications to be directly observed and analyzed. A
symmetry-based experimental design further reduces the influence of
confounding factors, allowing preferences associated with different representation types to be compared systematically.

We evaluate the framework on an executable mathematical benchmark consisting
of 550 conflict instances spanning 11 function families, comparing four
representation types: pure natural language, formal language, naturalized
formal language, and input--output examples. The results show that model
behavior under conflict is systematic rather than random, exhibiting a
consistent preference ordering of
\( \text{Formal} \approx \text{Naturalized Formal} >
\text{Pure Natural Language} > \text{Input--Output Examples} \).
We further find that the influence of examples depends on both model
capability and the underlying function family.

Furthermore, we extend the framework to heterogeneous specification conflicts
in Boolean algebra, code generation, and the clinical domain, demonstrating
its applicability across different tasks and specification forms. These
experiments show that the framework can characterize model preferences under
diverse conflict settings while preserving attributable preference
measurement, providing a unified experimental framework for analyzing conflict
resolution behavior in large language models across different domains.
\end{abstract}

\section{Introduction}

Large language models (LLMs) increasingly operate in settings where multiple
pieces of contextual information jointly specify how a task should be
performed~\cite{lewis2020retrieval, yang2018hotpotqa, feldman2019multi}.
These specifications may take diverse forms, including natural language
instructions, examples, formal constraints, and executable tests. In practice,
such specifications may be incomplete, inconsistent, or directly conflicting
~\cite{hou2024wikicontradict, zhang2025leveraging, liu2025conflicts}.
When two specifications support incompatible behaviors, the model must
implicitly determine which specification governs its final response.
Understanding this selection behavior is important because it reveals the
implicit hierarchy that models assign to different forms of specification and
is essential for designing reliable LLM-based systems.

Prior work has studied how LLMs detect, reconcile, or respond to conflicting
information in settings such as retrieval-augmented generation, fact-checking,
and multi-context question answering
~\cite{ge2025resolving, li2025bordarag, zeng2025towards,
hou2024wikicontradict, kurfali2025conflicting, lee2025magic}.
Other studies have examined how models select among competing evidence or
information sources~\cite{schuster-etal-2026-whose}. However, existing studies
have primarily focused on conflicts among information sources, while how
models prioritize different forms of task specification remains less
understood. In particular, it remains unclear whether LLMs exhibit systematic
preferences when the same underlying task is expressed through different
representations, how these preferences vary across models, and whether they
generalize beyond controlled settings.

To answer these questions, we introduce a controlled, symmetry-based framework
for measuring model preference under conflicting specifications. We represent a
task through competing executable mappings that produce different outputs on a
discriminative query. Because each specification supports a distinct candidate
output, the model's response can be directly attributed to one mapping, the
other mapping, or neither. We further construct symmetry-complete
configurations that counterbalance presentation order and the assignment of
representation types to candidate mappings. This turns conflict resolution
into an attributable specification-selection problem while reducing the
influence of non-target factors.

We instantiate the framework in a controlled mathematical benchmark containing
550 conflict instances across 11 function families and compare four
representation types: pure natural language, formal language, naturalized
formal language, and input--output examples. Our experiments reveal systematic
representation preferences across models, with Formal and Naturalized Formal
specifications consistently preferred over Pure Natural Language and
Input--Output Examples. We further show that naturalized formal specifications
behave much more like formal specifications than like an intermediate
representation, while the influence of examples varies across models and task
families.

Finally, we extend the framework to heterogeneous specification conflicts in
Boolean algebra, code generation, and clinical rule settings. These cases
demonstrate that the framework can measure attributable specification
preferences beyond the controlled mathematical benchmark. Overall, this work
establishes conflict resolution in LLMs as a controlled, executable, and
symmetry-counterbalanced measurement problem, enabling systematic analysis of
how models prioritize different forms of specification.

\section{Background and Related Work}

Prior studies have shown that conflicts are widespread across different types 
of contextual information and can affect LLM reliability. Liu and Roth analyze 
conflicts in natural texts, human annotations, and model interactions, while 
Xu et al. provide a systematic categorization of knowledge conflicts in LLMs. 
Zhou et al. further show that models may exhibit preferences among competing 
information sources, suggesting that conflict resolution depends on the 
properties of conflicting contexts and sources 
\cite{liu2025conflicts,xu2024knowledge,zhou2024establishing}.

Recent benchmarks have developed controlled evaluations of conflict 
resolution. WikiContradict studies whether models can recognize and 
synthesize contradictory but equally credible Wikipedia evidence 
\cite{hou2024wikicontradict}, while MAGIC introduces structured 
knowledge-graph-based conflicts for analyzing conflict detection and 
localization \cite{lee2025magic}. These works primarily focus on whether 
models can identify, localize, or reconcile conflicting information.

Beyond conflict resolution, recent work has examined preference among 
competing information sources. Wan et al. show through the CONFLICTINGQA 
benchmark that LLM decisions can be influenced by properties of competing 
evidence, which may not align with human judgments of credibility 
\cite{wan2024evidence}. Schuster et al. further study source preferences under 
inter-context knowledge conflicts and demonstrate systematic preferences 
toward different source types \cite{schuster-etal-2026-whose}. 

Complementary to these studies, Parasaram et al. show that different 
organizations and combinations of contextual information can lead to 
different LLM behaviors in software engineering scenarios 
\cite{parasaram2024fact}, suggesting that the representation and organization 
of task-relevant information can influence model decisions.

Our work studies a complementary but different dimension of preference. Whereas previous studies mainly vary source reliability, attribution, or context organization, we investigate how models choose between incompatible task specifications expressed through different representation types. Each specification corresponds to an executable candidate mapping, enabling model responses to be directly attributed to competing specifications and allowing preference to be measured under controlled conflicts.

\section{Method}

We propose a framework for measuring model preference under conflicting
specifications. The framework represents conflicts as competing executable
mappings and uses symmetry-complete counterbalancing to attribute model
outputs while reducing effects from non-target factors such as presentation
order and mapping assignment.

\subsection{A Symmetry-Based Framework for Identifiable Preference Attribution}

To characterize LLM preferences under conflicting specifications, we formulate a framework that represents model behavior under multiple competing specifications as a selection process over candidate mappings.

We begin by defining a set of conflicting specifications, each of which induces a different output for the same input. In this way, conflict is formalized as an inconsistency relation among candidate mappings.

Model outputs are then interpreted as the outcomes of this selection process. Since each specification determines a concrete output for the queried input, the model prediction can be directly attributed to one specification, turning an implicit decision into an observable variable.

Building on this formulation, measuring model preference requires handling two types of interfering factors. The first consists of factors that cannot be disentangled within a single configuration, such as presentation order, whose influence is coupled with other factors. To address this, we construct symmetric configurations so that the effect of such factors appears with opposite directions across configurations and cancels out in the aggregate.

The second consists of factors that cannot be precisely controlled, such as subtle differences in how content is expressed across specifications. Prior work has shown that even meaning-preserving changes in prompt formatting can substantially affect model behavior \cite{sclar2024quantifying}. Such variation may introduce unintended asymmetries and thereby affect selection behavior. To mitigate this source of bias, we construct specifications that are symmetric in structure and matched in informational content, while avoiding fixed surface templates that could introduce systematic lexical cues.

Symmetry operates at both the configuration and specification levels, reducing
uncontrolled asymmetries and enabling preference estimation through aggregation
over symmetry-complete configurations.

\subsection{Problem Formulation}

Let $\mathcal{X}$ denote the input space and $\mathcal{Y}$ the output space. A task is defined by a mapping
\[
f: \mathcal{X} \to \mathcal{Y}.
\]
A specification $s$ provides a description of the task and induces a candidate mapping
\[
f_s: \mathcal{X} \to \mathcal{Y}.
\]

We consider a set of specifications $\{s_1, s_2, \dots, s_k\}$, and assume that the corresponding candidate mappings produce distinct and attributable outputs for the query inputs under consideration.

Given a prompt containing a subset of these specifications and a query input $x$, the model produces a prediction $\hat{y}$. Since each specification defines a concrete output for the same input, the model prediction can be interpreted as selecting one of the candidate mappings.

Our objective is to characterize the model's selection behavior under such conflicts, i.e., which specification the model follows under different conditions.

\subsection{Preference Attribution}
\label{sec:pref_attr}

Under a fixed configuration, attribution yields a conditional quantity that depends on both the target factor and other factors in the setup. Rather than interpreting this quantity directly, we estimate preference by aggregating over a symmetry-complete set of configurations.

Specifically, let $\mathcal{U}$ denote the set of configurations generated by symmetric transformations over non-target factors. We define the preference for a target factor value $t$ as
\[
P(t)
=
\mathbb{E}_{u \in \mathcal{U}}
\;
\mathbb{P}\big(
\text{select } t \mid t, u
\big).
\]
where ``select $t$'' denotes selecting a specification with factor value $t$.

By construction, symmetry operates at two levels: configuration-level symmetry offsets factors such as order that cannot be independently isolated, while specification-level symmetry reduces unintended asymmetries in information content. The resulting $P(t)$ provides a preference estimate after marginalizing over
symmetry-controlled configurations, reducing sensitivity to
configuration-specific artifacts.

\section{Experimental Setup}

\begin{figure}[t]
    \centering
    \includegraphics[width=1\linewidth]{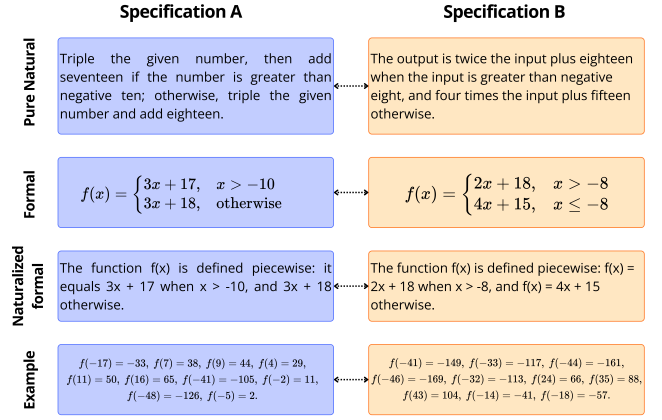}
    \caption{
Example conflict instance represented in four specification formats:
pure natural language (NL), formal specification (Form), naturalized formal
specification (NatF), and input--output examples (Ex).
    }
    \label{fig:f1_example}
\end{figure}

The framework constructs conflicts between two executable mappings, \(f_1\) and
\(f_2\), represented through four specification formats: pure natural
language, formal specifications, naturalized formal specifications, and
input--output examples, as shown in Figure \ref{fig:f1_example}. Specification order and mapping assignment are
counterbalanced, allowing model outputs to be attributed to \(f_1\), \(f_2\),
or \emph{other}.

\subsection{Mathematical Conflict Benchmark}
\label{sec:math-benchmark}

We construct a mathematical benchmark for measuring model behaviour under
conflicting specifications. Mathematics provides executable semantic ground
truth: each task is defined by a function, allowing model outputs to be
directly compared against competing candidate mappings.

The benchmark contains 550 conflict instances from 11 parameterized function
families, with 50 independently generated instances per family. The families
cover diverse executable mathematical tasks, including algebraic functions,
conditional rules, sequences, polynomial queries, graph problems, and matrix
operations, with parameters sampled from bounded ranges to avoid degenerate
cases.

For every base mapping \(f_1\), we construct a conflicting mapping \(f_2\)
through family-specific, schema-preserving perturbations. These perturbations
maintain the original task structure while inducing different executable
outputs. We then generate a discriminative query \(q^\star\) satisfying
\[
f_1(q^\star)\neq f_2(q^\star).
\]

For each conflict pair, the dataset stores \(q^\star\), the corresponding
outputs, and executable implementations of both mappings, enabling automatic
attribution during evaluation. For example-based representations, we
additionally provide input--output examples sampled from each mapping while
excluding \(q^\star\), ensuring that the discriminative query cannot be
answered by direct retrieval.

For each mapping, DeepSeek-V4-Pro \cite{deepseekv4} generates three specification realizations:
natural language, formal, and naturalized formal specifications. The model is
used only for realizing predefined executable semantics; all mappings, queries,
outputs, and labels are determined statically.

\subsection{Experimental Protocol}
\label{sec:experimental-protocol}

We evaluate how models resolve conflicts between four representation types:
pure natural-language specifications, formal specifications, naturalized
formal specifications, and input--output examples. We consider all six
unordered representation pairs:
pure natural language versus formal,
pure natural language versus naturalized formal,
pure natural language versus examples,
formal versus naturalized formal,
formal versus examples, and
naturalized formal versus examples.

Each trial is constructed from one conflict instance containing two conflicting
executable mappings, \(f_1\) and \(f_2\), and a discriminative query
\(q^\star\). The two presented specifications describe different mappings: one
realizes \(f_1\), while the other realizes \(f_2\). For each representation pair
\(A,B\), where \(A\) and \(B\) denote the two representation types being
compared and \(A(f_i)\) denotes mapping \(f_i\) expressed using representation
type \(A\), we generate four counterbalanced trials per conflict instance:
\[
\begin{array}{cc}
(A(f_1), B(f_2)) &
(B(f_1), A(f_2)) \\[3pt]
(A(f_2), B(f_1)) &
(B(f_2), A(f_1))
\end{array}
\]
This design counterbalances both the presentation order of the two
specifications and the assignment of representation types to the underlying
mappings.

Example-based sources contain coherent input--output examples generated from
their associated executable mapping. Most instances provide ten examples per
mapping; graph degree queries provide four examples because of the smaller
graph representation. The examples do not include the discriminative query
\(q^\star\).

\begin{figure}[t]
    \centering
    \includegraphics[width=1\linewidth]{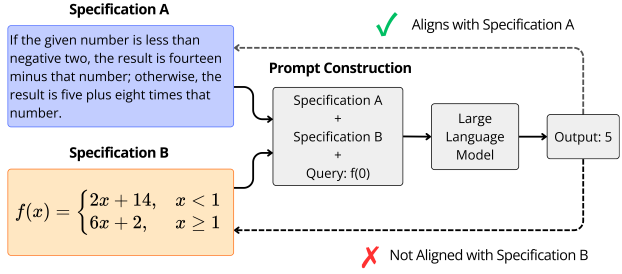}
\caption{Example prompt construction and output attribution procedure.
Responses are attributed to \(f_1\), \(f_2\), or \emph{other} by comparing model
outputs with the stored candidate outputs.}
    \label{fig:prompt_example}
\end{figure}

In every trial, the model receives the two conflicting specifications and is
asked to answer the query \(q^\star\), as illustrated in
Figure~\ref{fig:prompt_example}. Models are instructed to provide a clearly
identifiable final answer, which is then extracted and normalized for
attribution. We compare the extracted value against the stored executable
outputs \(f_1(q^\star)\) and \(f_2(q^\star)\). A response matching the first
output is attributed to \(f_1\), a response matching the second is attributed
to \(f_2\), and a response matching neither is assigned to the \emph{other}
category.

\subsection{Generalizing the Framework to Heterogeneous Specification Conflicts}
\label{sec:heterogeneous-cases}

We further evaluate the framework on Boolean algebra, code generation, and clinical rule conflicts; construction and validation details are provided in the supplementary material.

\subsection{Research Questions}

To investigate this question, we study the following research questions:

\begin{itemize}
    \item \textbf{RQ1.} When different representations of the same task support
    incompatible mappings, do language models exhibit systematic preferences
    for particular representation types?

    \item \textbf{RQ2.} How do representation preferences vary across
    different language models?

    \item \textbf{RQ3.} How does the framework extend to heterogeneous
    specification conflicts beyond the controlled mathematical benchmark?
\end{itemize}

\section{Results}

We organize the results around the three research questions.
We first evaluate representation preferences in the controlled
mathematical benchmark (RQ1), analyze variation across models
(RQ2), and finally examine heterogeneous specification conflicts
beyond the controlled benchmark (RQ3).

\subsection{Pairwise Representation Preferences}
\label{sec:pairwise-preferences}

\begin{figure*}[t]
    \centering
    \includegraphics[width=\textwidth]{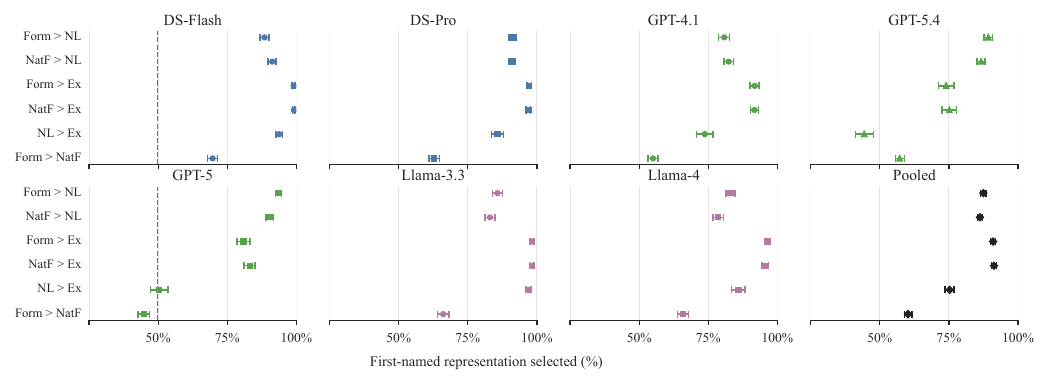}
    \caption{
    Pairwise representation preferences across evaluated models.
    Each point shows the proportion of attributable responses selecting the
    first-named representation in the comparison label. The dashed line marks
    50\%. Error bars show clustered-bootstrap 95\% confidence intervals over
    conflict instances. The final panel aggregates results across models.
    }
    \label{fig:pairwise-preferences}
\end{figure*}

Models consistently favor specifications that retain explicit mathematical
structure. As shown in Figure~\ref{fig:pairwise-preferences}, formal
specifications are selected over pure natural-language specifications in
\(87.45\%\) of attributable responses, and naturalized formal specifications
are selected over pure natural-language specifications in \(86.77\%\).
The gap is even larger when the alternative is a set of input--output
examples: formal specifications are selected over examples in \(93.04\%\) of
responses, while naturalized formal specifications are selected over examples
in \(93.27\%\). Pure natural-language specifications also beat examples
overall, but less decisively, with a selection rate of \(77.80\%\).

The main pattern is therefore clear:
\[
\begin{aligned}
\text{Formal} &\approx \text{Naturalized Formal} \\
&> \text{Pure Natural Language} \\
&> \text{Input--Output Examples}.
\end{aligned}
\]
This ordering is broadly reflected across models, although the relative
preference between pure natural language and examples varies across models.

Formal and naturalized formal specifications are the only pair without a
uniform direction. Overall, models select formal specifications in \(59.67\%\)
of their direct comparisons. However, DeepSeek-V4-Flash \cite{deepseekv4} favors formal
specifications strongly (\(69.67\%\)), whereas GPT-5-nano  \cite{openai2025gpt5} instead favors
naturalized formal specifications (\(55.24\%\)). Thus, models agree much more
strongly on the value of preserving mathematical structure than on whether
that structure should be expressed as pure notation or embedded in natural
language.

\subsection{When Do Examples Become Competitive?}
\label{sec:example-use}

\begin{figure*}[t]
    \centering
    \includegraphics[width=\textwidth]{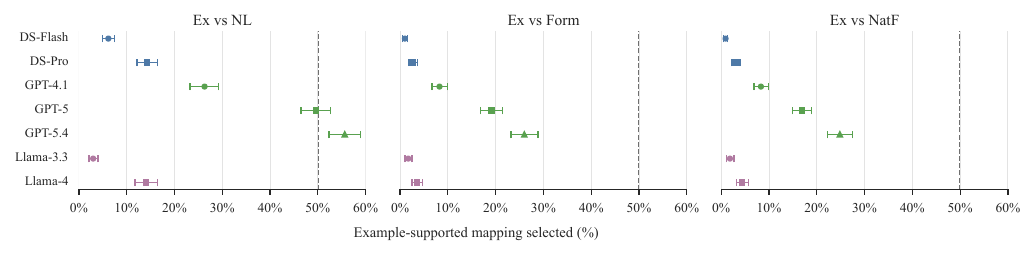}
    \caption{
Selection of example-supported mappings across models.
Each point shows the proportion of attributable responses selecting the
mapping specified by input--output examples under conflicts with different
representation types.
}
    \label{fig:example-use-across-models}
\end{figure*}

Figure~\ref{fig:example-use-across-models} summarizes example selection across
models. The influence of coherent input--output examples varies substantially
across models. This variation is most pronounced when examples compete with
pure natural-language specifications. In this condition, GPT-5.4-nano \cite{openai2026gpt54nano} selects
the example-supported mapping in \(55.64\%\) of attributable responses, whereas Llama-3.3-70B \cite{llama3} selects it in only
\(2.98\%\). GPT-5-nano is approximately balanced in this condition, selecting
the example-supported mapping in \(49.63\%\) of attributable responses. The remaining models fall between these
extremes, with example-selection rates ranging from \(6.16\%\) to \(26.29\%\).
Thus, example selection varies substantially across models when examples
compete with pure natural-language specifications.

Examples are substantially less selected when the opposing specification
preserves explicit mathematical structure. Against formal specifications,
example selection ranges from \(1.00\%\) for DeepSeek-V4-Flash to
\(26.00\%\) for GPT-5.4-nano. The corresponding range against naturalized
formal specifications is \(0.91\%\) to \(24.84\%\). These results show that
model variation in example selection does not change the overall preference
pattern observed in the main benchmark: examples are selected less frequently
when the competing specification exposes explicit mathematical structure.
 
Within-family comparisons reveal systematic differences between model variants.
Table~\ref{tab:within-family-example-diffs} summarizes the differences between
model variants within the same model family. Across all evaluated families,
higher-capability variants tend to select example-supported mappings more
frequently, although the magnitude of the difference varies across
representation conflicts.

\begin{table}[t]
\centering
\small
\begin{tabular}{lccc}
\toprule
Model family 
& NL vs. Ex 
& Formal vs. Ex 
& NatF vs. Ex \\
\midrule
DeepSeek
& +8.10 & +1.74 & +2.17 \\
GPT
& +29.35 & +17.75 & +16.52 \\
Llama
& +11.12 & +1.83 & +2.54 \\
\bottomrule
\end{tabular}
\caption{Maximum within-family differences in example selection.
Values indicate percentage-point differences among model variants with
different capability levels within the same model family.}
\label{tab:within-family-example-diffs}
\end{table}

\subsection{Robustness Checks}
\label{sec:robustness}

Presentation order has a secondary effect, but it does not explain the
representation preferences reported above. Across five of the six
representation pairs, the pooled probability of selecting the specification
presented second ranges from \(51.30\%\) to \(54.53\%\), which is small relative
to the larger differences associated with representation type. The main
exception is the direct comparison between formal and naturalized formal
specifications, where the second-presented specification is selected in
\(67.86\%\) of attributable responses. This suggests that position is more
influential when competing specifications have similar overall strength.
Additional presentation-order analyses are provided in the supplementary material.

We further examine whether the observed preferences could be affected by
properties of the competing mappings rather than their representation forms.
Mapping identity is close to balanced: the minimum and maximum rates of
attributable responses selecting \(f_2\) across models are \(50.64\%\) and
\(51.66\%\), respectively. Responses assigned to \emph{other} account for
\(0.67\%\) to \(5.10\%\) of all trials and are concentrated in the
pure-natural-language versus examples condition, where the pooled rate is
\(6.36\%\), compared with \(1.34\%\) to \(2.02\%\) for the remaining
comparisons. We therefore report pairwise specification-selection rates over
attributable responses throughout.

\subsection{Heterogeneous Specification Conflicts}
\label{sec:heterogeneous-results}

We next evaluate the framework on three heterogeneous specification-conflict
settings: boolean algebra, code generation, and clinical rule conflicts.
These settings differ in their specification forms and output spaces,
allowing us to examine whether attributable preference measurement can be
maintained beyond the controlled mathematical benchmark.

\subsubsection{Boolean Algebra}
\begin{figure}
    \centering
    \includegraphics[width=1\linewidth]{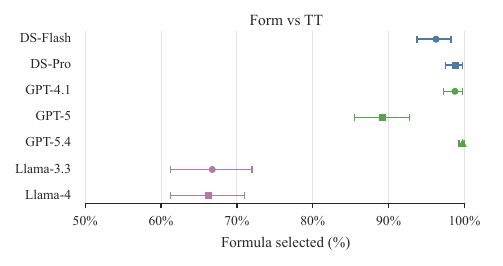}
    \caption{
    Results on the Boolean Algebra dataset, showing model preferences between
    Boolean formulas (Form) and truth tables (TT).
    }
    \label{fig:boolean_algebra}
\end{figure}
We constructed boolean-algebra conflict instances, as in the main experiments, the detailed construction is illustrated in the supplementary material. Each instance contained two boolean expressions that differed through a localized modification. One expression was represented by a symbolic boolean formula, while the other was represented by its truth table.

Across from selected models, as shown in Figure \ref{fig:boolean_algebra}, models exhibited consistent preference for boolean formula: the formula-selection rate was 87.96\%. Also, for the non-Llama models, presentation order and mapping identity remained approximately balanced. In contrast, the two Llama-family models exhibited a marked second-position preference, 70.25\% and 66.75\% of responses, but still follows the result observed in \emph{Robustness Checks} section.

This preference is particularly notable because the unambiguous, exhaustive truth table specification allowed direct lookup for discriminative queries, whereas the formula required evaluation. If models prioritized computational convenience or explicit completeness, the truth table would be highly competitive. However, result suggests that, under conflict, models strongly favor compact and compositionally structured symbolic representations over exhaustive tabular ones.  This finding extends the main benchmark result beyond the original representation types and shows that expression preference persists.

Additional comparisons involving gate-netlist are reported in the supplementary material, showing that gate netlists are similarly preferred over truth tables, while formula vs gate-netlist comparisons remain model-dependent.

\subsubsection{Code Generation}

We next apply our framework to code generation, where natural-language
documentation and executable tests specify incompatible behaviors. From MBPP
\cite{austin2021program}, we construct 67 verified conflict pairs that preserve
the original function signature and input domain but differ in one semantic
slot, with one behavior grounded in the original task and the other forming a
minimal alternative. For each pair, we combine documentation for one behavior
with visible tests for the other in four counterbalanced trials. Generated
implementations are evaluated against validated oracle suites and attributed to
the documentation-supported behavior, the test-supported behavior, or
\emph{other}. Further details are provided in
the supplementary material.

\begin{figure}[t]
    \centering
    \includegraphics[width=1\linewidth]{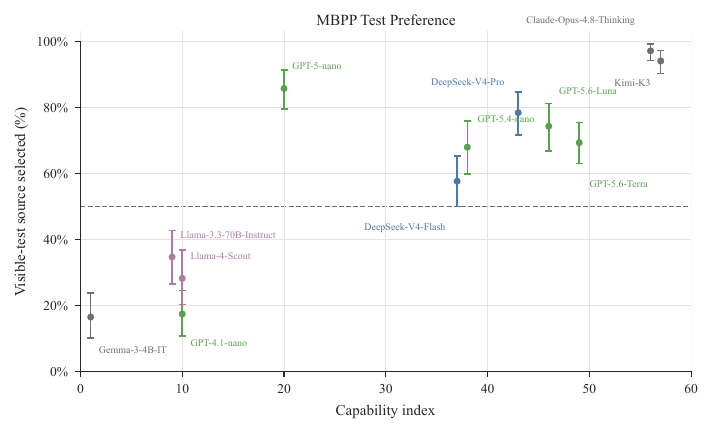}
    \caption{Test preference across models with different capability scores. Model capability is measured using the Artificial Analysis Intelligence Index v4.1~\cite{artificial_analysis_intelligence_2026,artificial_analysis_models_2026}}
    \label{fig:code-conflicts}
\end{figure}

As shown in Figure~\ref{fig:code-conflicts}, specification preference varies sharply across models. Test-selection rates range from 16.5\% for Gemma-3-4B-IT \cite{gemma3} and 17.4\% for GPT-4.1-nano \cite{openai2025gpt41} to 94.0\% for Kimi-K3 \cite{kimik3} and 97.1\% for Claude Opus 4.8 \cite{anthropic2026opus48}. Although test preference does not increase monotonically with the external capability score, the models exhibit a broad capability-associated separation: lower-scoring models generally favor documentation, whereas the highest-scoring models strongly favor tests.

\subsubsection{Clinical Rule Conflicts}
\label{sec:clinical-case}

We extend the framework to authentic clinical rules drawn from public-domain
openFDA labels~\citep{fda_openfda_labels_2026}. We construct 166 conflicts over
renal dosing thresholds and pediatric minimum-age indications, pairing each genuine threshold with a
schema-preserving perturbation. Perturbations are applied in both directions,
making the genuine rule more restrictive in roughly half of the instances and
allowing restrictiveness to be separated from correctness. Rendering both rules
in all four representation types under the symmetry design yields $7{,}968$
counterbalanced trials across nine models from five families.
Appendix details the construction, coverage, and attribution
procedures. These rules are transcribed solely to measure model selection
behavior and do not constitute medical advice.

\begin{figure}[t]
    \centering
    \includegraphics[width=\linewidth]{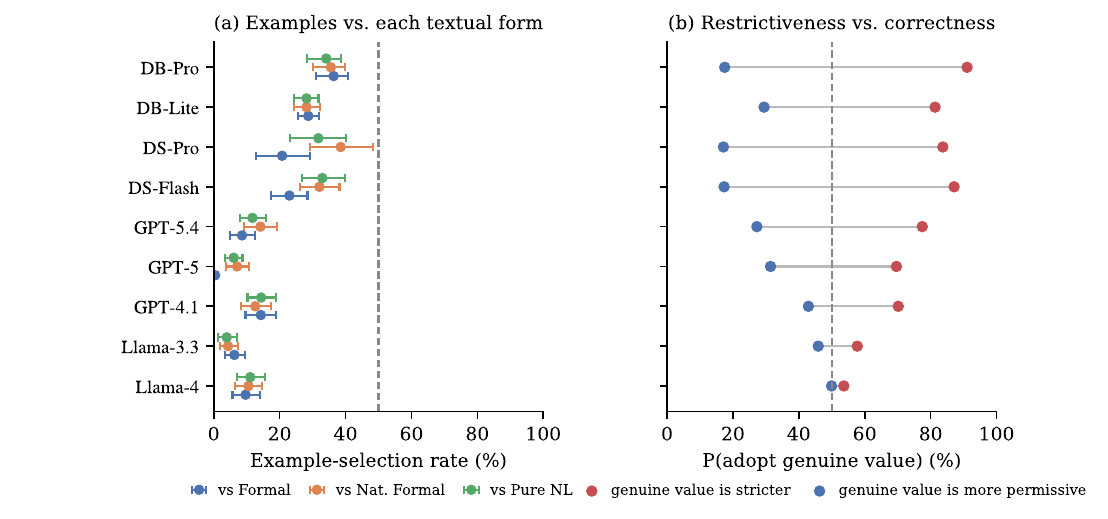}
    \caption{Clinical rule conflicts. \textbf{(a)} Example-selection rates
    against each textual form; all clustered-bootstrap intervals lie below
    $50\%$. \textbf{(b)} Genuine-value selection conditioned on whether the
    genuine rule is stricter or more permissive.}
    \label{fig:medical-combined}
\end{figure}

Across all $27$ model--text comparisons, examples are the least competitive
specification: selection rates range from $0.4\%$ to $38.5\%$, with every
clustered-bootstrap interval excluding $50\%$
(Figure~\ref{fig:medical-combined}a). The three textual forms show no stable
ordering, so the mathematical ordering does not transfer to this domain.

All nine models prefer the more restrictive rule ($52\%$--$87\%$). Genuine-value
selection remains near chance overall ($51\%$--$57\%$), but splits to
$70\%$--$91\%$ when the genuine rule is stricter and $17\%$--$43\%$ when it is
more permissive for models with the conservative tilt
(Figure~\ref{fig:medical-combined}b). Given the bidirectional perturbations,
this split tracks restrictiveness rather than label correctness, revealing a
domain-specific selection bias rather than a representation preference.

\label{sec:real-world-results}

\section{Discussion}

Our results show that models exhibit systematic preferences among
representation forms under conflicting specifications. Formal and naturalized
formal specifications are selected more frequently than less structured
alternatives, suggesting that explicit structure facilitates conflict
resolution by directly exposing key constraints of the underlying mapping.
This may also explain why naturalized formal specifications behave similarly
to formal ones despite their more natural-language style. Pure
natural-language and example-based specifications require greater inference
from descriptions or finite observations, but examples are not inherently
weaker. Their influence depends on model capability and task structure:
higher-capability variants select example-supported mappings more often, and
examples are more competitive when the underlying rule is easier to recover
from demonstrations, particularly against pure natural-language
specifications. This interpretation is consistent with Wei
et al.~\cite{wei2023larger}, who find that learning input--label mappings from
semantically unrelated exemplars improves with model scale. The heterogeneous
settings further show that these preferences are context-dependent: Boolean
algebra favors compact structured representations, whereas clinical conflicts
reveal additional domain-specific influences beyond representation form.

In code generation, conflicting software artifacts require models to select
among competing intended behaviors rather than simply synthesize programs from
a single specification. This reflects a latent specification-selection policy whose effects are not inherently desirable or undesirable. Tests may provide precise executable specifications, but they may also be incomplete, stale, or erroneous. Reliable systems should therefore resolve conflicts contextually rather than universally prioritizing tests or documentation.

Our main experiments characterize conflict resolution when models are
required to produce a task answer, but specification selection is not the only
possible response to conflicting specifications. In an additional diagnostic
experiment, we minimally modified the original prompt to allow an explicit
\emph{uncertain} response while keeping the conflict instances and experimental
conditions unchanged. We evaluated this setting on two representative models:
DeepSeek-V4 and GPT-5-nano selected this option in approximately \(6\%\) and
\(17\%\) of trials, respectively. These results show that explicit abstention
can constitute an additional conflict-handling strategy, and suggest that the
tendency to defer resolution may vary across models. Full experimental details
are provided in the supplementary material.

Our controlled framework isolates representation preferences in conflicting
specifications, but several limitations remain. Real-world conflicts are often
more complex and may involve incomplete, ambiguous, or partially overlapping
specifications. Moreover, different representations cannot be fully separated
from accompanying factors such as description length, information density, and
level of detail, which may jointly contribute to the observed preferences.
Finally, our framework relies on executable outputs to infer specification
selection by comparing model responses with candidate specifications. Therefore,
it is most naturally applicable to tasks with explicit behavioral mappings.

\section{Conclusion}
We introduced a symmetry-based framework that represents conflicting
specifications as competing executable mappings, enabling model responses to
be attributed under counterbalanced configurations. Across the mathematical
benchmark, models systematically favored formal and naturalized formal
specifications, while the influence of examples depended on model capability
and task structure. The heterogeneous experiments further showed that this
ordering is not universal: preferences varied across software and clinical
settings, where domain-specific factors also shaped selection. Together, these
results characterize LLM conflict resolution as structured but
context-dependent specification selection and provide a controlled approach
for measuring such behavior across tasks.

\newpage
\bibliography{references}

\appendix

\section{Mathematical Benchmark Construction}
\label{app:benchmark-construction}

This appendix describes the construction of the mathematical
conflict-specification benchmark. The current benchmark contains
550 conflict instances from 11 function families, with 50 independently
generated instances per family. Each instance consists of two executable
candidate mappings, denoted by \(f_1\) and \(f_2\), that share the same task
family and input--output schema but differ in selected metadata fields.

The benchmark covers algebra, calculus, graph reasoning, conditional rules,
sequence reasoning, and matrix computation. Rather than treating all conflicts
as uniform numerical perturbations, we construct family-specific alternative
mappings that preserve the natural structure of each underlying task. The goal
is to make the two candidate functions minimally different while remaining
meaningfully distinguishable: each pair retains the same function family,
input--output schema, and overall mathematical form, but differs on at least
one valid query. This design allows the benchmark to vary the underlying
mathematical content without creating artificial or malformed tasks, while
preserving executable semantics and automatic output attribution.

\subsection{Function Families}
\label{app:function-families}

Table~\ref{tab:benchmark-families} summarizes the function families,
their input--output schemas, and the corresponding conflict constructions.
Metadata are sampled from bounded ranges, subject to family-specific
validity constraints described below.

\begin{table*}[h]
\centering
\small
\setlength{\tabcolsep}{5pt}
\renewcommand{\arraystretch}{1.12}

\begin{tabular}{L{0.25\textwidth} L{0.23\textwidth} L{0.42\textwidth}}
\toprule
Function family & Input / output & Conflict construction \\
\midrule

\multicolumn{3}{l}{\textit{Algebra}} \\
\addlinespace[1pt]

Linear functions
& Integer \(\rightarrow\) integer
& Perturb the slope, the intercept, or both. \\

Quadratic functions
& Integer \(\rightarrow\) integer
& Perturb the leading coefficient, the linear coefficient, or both; keep the constant term fixed. \\

Rational functions
& Valid integer input \(\rightarrow\) reduced fraction
& Perturb one or both numerator coefficients; keep the denominator fixed. \\

\addlinespace[2pt]
\multicolumn{3}{l}{\textit{Conditional rules}} \\
\addlinespace[1pt]

Threshold-conditioned functions
& Integer \(\rightarrow\) integer
& Perturb the threshold and both branch mappings; optionally change the comparison operator. \\

Divisibility-conditioned functions
& Integer \(\rightarrow\) integer
& Perturb the divisor and both branch mappings. \\

\addlinespace[2pt]
\multicolumn{3}{l}{\textit{Sequences and calculus}} \\
\addlinespace[1pt]

Linear recurrence sequences
& Nonnegative integer index \(\rightarrow\) integer
& Perturb a nonempty subset of the initial value, multiplicative coefficient, and additive coefficient. \\

Polynomial derivative queries
& Integer \(\rightarrow\) integer
& Perturb exactly one polynomial coefficient. \\

Polynomial integral queries
& Integer interval \(\rightarrow\) integer or reduced fraction
& Perturb exactly one polynomial coefficient. \\

\addlinespace[2pt]
\multicolumn{3}{l}{\textit{Graph reasoning}} \\
\addlinespace[1pt]

Graph degree queries
& Vertex \(\rightarrow\) integer
& Add or remove exactly one edge while preserving connectivity. \\

Shortest-path queries
& Source--target pair \(\rightarrow\) integer distance
& Add or remove exactly one edge while preserving connectivity. \\

\addlinespace[2pt]
\multicolumn{3}{l}{\textit{Matrix computation}} \\
\addlinespace[1pt]

Shifted matrix determinants
& Integer matrix \(\rightarrow\) integer
& Perturb exactly one entry of the shift matrix. \\

\bottomrule
\end{tabular}

\caption{Function families in the mathematical conflict-specification benchmark. Each family contains 50 independently generated conflict instances. All instances are filtered to preserve family-specific validity constraints and to admit at least one discriminative query.}
\label{tab:benchmark-families}
\end{table*}

Across all families, we reject instances that are undefined, degenerate,
or not distinguishable under the two candidate mappings. In particular,
linear functions require nonzero slopes and quadratic functions require
nonzero leading coefficients; rational functions must have nonzero
denominator parameters and remain defined on all query and example inputs;
threshold- and divisibility-conditioned functions require valid branch
parameters and nonzero branch slopes; recurrence queries use bounded indices
and valid recurrence coefficients; graph instances must remain connected,
contain no self-loops, and yield different values for the selected query;
definite integrals require valid ordered bounds; and matrix instances are
retained only when their entries and determinant outputs remain bounded and
easily representable. For every family, we additionally require that the two
candidate mappings differ on at least one valid query. The complete
family-specific metadata schemas, sampling ranges, and generation constraints
are released with the benchmark.

For each retained conflict pair, we generate a discriminative query
\(q^\star\) such that
\[
f_1(q^\star) \neq f_2(q^\star).
\]
The dataset stores \(q^\star\), the corresponding outputs
\(f_1(q^\star)\) and \(f_2(q^\star)\), and executable implementations of
both mappings. It also stores up to ten input--output examples for each
mapping where applicable; the discriminative query is excluded from these
example sets. Thus, model responses can be attributed automatically to
\(f_1\), \(f_2\), or \emph{other} by comparing the normalized final answer
with the two stored outputs. Table~\ref{tab:benchmark-examples} presents
representative conflict instances from different families.

\begin{table*}[t]
\centering
\small
\setlength{\tabcolsep}{5pt}
\renewcommand{\arraystretch}{1.08}

\begin{tabular}{L{0.16\textwidth} L{0.36\textwidth} L{0.18\textwidth} L{0.20\textwidth}}
\toprule
Family & Conflict pair & Query & Outputs \\
\midrule

Linear
& \(a=9, b=-23 \;\rightarrow\; a=9, b=-27\)
& \(x=-15\)
& \(-158\) vs.\ \(-162\) \\

Quadratic
& \((a,b,c)=(-4,-11,-15) \;\rightarrow\; (-3,-2,-15)\)
& \(x=-25\)
& \(-2240\) vs.\ \(-1840\) \\

Rational
& \((a,b,c,d)=(3,11,7,10) \;\rightarrow\; (7,14,7,10)\)
& \(x=-6\)
& \(7/32\) vs.\ \(7/8\) \\

Degree query
& Remove edge \((0,1)\)
& \(v=0\)
& \(2\) vs.\ \(1\) \\

Shortest path
& Add edge \((0,4)\)
& \((1,0)\)
& \(3\) vs.\ \(2\) \\

Derivative
& Coefficients \([4,-5,3] \;\rightarrow\; [4,-5,1]\)
& \(t=7\)
& \(37\) vs.\ \(9\) \\

Definite integral
& Coefficients \([1,-4,-3] \;\rightarrow\; [1,-4,-1]\)
& \([-4,5]\)
& \(-198\) vs.\ \(-72\) \\

Matrix determinant
& Shift \(\begin{bmatrix}2 & -3\\-1 & 0\end{bmatrix}
\;\rightarrow\;
\begin{bmatrix}2 & -1\\-1 & 0\end{bmatrix}\)
& \(\begin{bmatrix}-3 & -3\\3 & 2\end{bmatrix}\)
& \(10\) vs.\ \(6\) \\

\bottomrule
\end{tabular}

\caption{Representative conflict instances. Each row shows a base mapping and its schema-preserving alternative, together with a query on which their executable outputs differ.}
\label{tab:benchmark-examples}
\end{table*}
\section{Specification Generation}
\label{app:specification-generation}

We use DeepSeek-V4-Pro to generate surface realizations of statically
defined executable mappings. Each realization corresponds to the same
underlying mapping, while differing only in representation style. The
generator is not given source code, structured metadata, discriminative
queries, candidate outputs, or the opposing mapping. Instead, each call
receives only a rendered formula or task-description string corresponding to
one executable mapping.

For quality control, we developed an HTML-based visualization tool that
displayed each generated specification alongside its corresponding executable
mapping and the associated conflict instance. All authors participated in
manual spot-checking across the three representation styles and the 11
function families. The inspection focused on semantic faithfulness to the
assigned mapping and on whether the generated realization introduced any
unintended semantic change beyond the predefined difference between the two
mappings.

For every conflict instance, \(f_1\) and \(f_2\) are processed
independently. For each mapping, we generate three specification styles:
pure natural language, formal specification, and naturalized formal
specification. Thus, each conflict pair yields six independently generated
specifications. Pure natural-language specifications prohibit notation and
symbolic formulas; formal specifications request minimal \LaTeX{}
mathematical expressions without explanatory prose; and naturalized formal
specifications retain mathematical notation where useful while expressing
the mapping in natural language and remaining close to the rendered input.
The executable mappings, discriminative queries, candidate outputs, and
attribution labels are generated statically and do not depend on the
language model used for specification realization.

All specifications are generated with DeepSeek-V4-Pro using temperature
\(1\), with reasoning disabled through
\texttt{thinking.type = disabled}. For reproducibility, we release the
generation prompts, rendered inputs, generated specification texts, and
generation code with the benchmark.

\paragraph{Pure natural language prompt.}
\begin{quote}
\small
Write a pure natural-language definition of the
\texttt{\{content\_type\}} below. Do not use mathematical notation,
symbolic formulas, programming terminology, bullet points, or variable
assignments. Define the function behavior only; do not explain how standard
mathematical operations are computed. Do not include introductory
meta-commentary such as `Here is a description'.

\medskip
\texttt{\{content\_label\}}:

\texttt{\{content\}}
\end{quote}

\paragraph{Formal specification prompt.}
\begin{quote}
\small
Write a minimal formal mathematical specification of the
\texttt{\{content\_type\}} below. Output only clean \LaTeX{} math,
preferably one or two displayed equations. Do not add explanatory prose,
examples, boxes, bullet points, or repeated restatements. Do not expand
standard operators such as determinants, sums, binomial coefficients,
derivatives, integrals, powers, or matrix products. Do not output a complete
\LaTeX{} document: do not include \texttt{documentclass},
\texttt{usepackage}, \texttt{begin document}, or \texttt{end document}.

\medskip
\texttt{\{content\_label\}}:

\texttt{\{content\}}
\end{quote}

\paragraph{Naturalized formal specification prompt.}
\begin{quote}
\small
Write a natural-language description of the
\texttt{\{content\_type\}} below. You may use mathematical notation when it
makes the computation clearer; it does not need to be pure natural language.
Keep the description close to the given specification; do not simplify,
expand, rearrange, or decompose expressions into step-by-step computation.
Do not include examples.

\medskip
\texttt{\{content\_label\}}:

\texttt{\{content\}}
\end{quote}
\section{Boolean Algebra Dataset Construction}
\label{app:bool-construction}

The boolean algebra dataset is designed to study how language models resolve conflicts other than only mathematical functions or natural languages. Each instance consists of two boolean specifications, denoted as \texttt{spec1} and \texttt{spec2}, which define two distinct boolean expressions over four binary variables \(A,B,C,D\). Each expression is represented as a recursively constructed symbolic expression. Atomic expressions are variables, optionally preceded by negation, and compound expressions are formed by combining two subexpressions with one of the binary operators \(\land\), \(\lor\), or \(\oplus\). Boolean outputs are encoded as binary values in \(\{0,1\}\).

For each instance, the generator first samples a boolean expression with maximum recursive depth three. Variables are sampled from \(A,B,C,D\), with negation inserted probabilistically, and binary compositions are sampled from AND, OR, and XOR. To include cases where syntactic grouping is important, a subset of expressions is generated in explicitly bracketed three-variable forms. A conflicting counterpart is then constructed by applying one of three perturbation types. In an \emph{operator flip}, one binary operator is replaced by a different binary operator. In a \emph{negation flip}, a negation is either inserted into or removed from a selected subexpression. In a \emph{grouping shift}, the parenthesization of a three-variable expression is changed. For example, from \((a \circ_1 b) \circ_2 c\) to \(a \circ_1 (b \circ_2 c)\), with distinct operators used so that the regrouping can alter the induced Boolean function.

After constructing a candidate pair, the generator exhaustively evaluates both specifications on all possible truth assignments. Pairs that are semantically equivalent are rejected. For each accepted pair, one distinguishing input \(x^\ast\) is selected such that
\[
f_1(x^\ast) \ne f_2(x^\ast),
\]
and the two corresponding outputs are stored as well. This guarantees that every data instance contains an explicit conflict whose resolution determines the answer to the query.

Each boolean specification is rendered deterministically into multiple representation types. The \emph{formula} representation gives a LaTeX Boolean formula using standard logical symbols. The \emph{truth-table} representation enumerates all input assignments together with the corresponding values. The \emph{gate-netlist} representation expresses the same computation as a sequence of logic gates, using intermediate gate variables and a final output assignment. Unlike the natural-language specifications used in some other domains, the boolean representations are generated directly from the symbolic expression and do not require a language-model rewriting step.

The final boolean algebra dataset contains 100 conflict instances. The realized dataset includes 32 operator-flip conflicts, 33 negation-flip conflicts, and 35 grouping-shift conflicts. Each instance has a unique first specification, a distinguishing input, both conflicting outputs, and deterministic renderings of both specifications in the supported representation formats.

\section{Code-Generation Conflict Construction}
\label{app:code-generation}

This appendix describes the code-generation experiment summarized in
Section \emph{Heterogeneous Specification Conflicts} and Figure~6.  It instantiates the same measurement principle as
the mathematical benchmark in a software setting: two sources specify
incompatible executable mappings, a target model must produce one
implementation, and the implementation is attributed by executing it against
independently constructed oracles.  Here the competing source types are a
natural-language requirements document and a finite set of visible unit tests.

\subsection{End-to-end construction pipeline}

The canonical source is the sanitized MBPP file, which contains 427 Python
tasks.  Each record provides a task identifier, a natural-language prompt, a
reference implementation, and public tests (with setup and challenge-test
fields where available).  For the reported experiment, we sampled 200 source
tasks without replacement using a local pseudo-random  generator with seed 42.
The unit of sampling was an MBPP task, rather than a successfully generated
pair: a rejected task was not replaced by another draw.  After semantic and
execution-based filtering, 67 tasks yielded verified conflict pairs.  Each pair
then produced four counterbalanced target-model trials, for a total of
$67 \times 4 = 268$ trials per model.

The complete pipeline was:
\begin{enumerate}
    \item sample MBPP source records deterministically;
    \item construct an original-grounded specification $A$ and a minimally
          changed counterfactual specification $B$;
    \item apply host-side structural checks and an independent model-evaluator
          call;
    \item repair or restart rejected candidates, with at most three semantic
          generation attempts per source task;
    \item generate reference implementations for $A$ and $B$, generate shared
          test inputs, and derive all expected outputs by executing the
          reference implementations;
    \item retain only pairs that pass the complete deterministic cross-suite
          verification matrix; and
    \item cross source assignment with presentation order, query each target
          model once per trial, and attribute the returned implementation by
          hidden execution.
\end{enumerate}
Every intermediate prompt is versioned.  Dataset hashes, prompt versions,
sampling settings, model identifiers, test counts, and execution settings are
included in configuration fingerprints so that incompatible artifacts are not
silently reused.

\subsection{Constructing a one-slot semantic conflict}

For a sampled record, the original MBPP prompt, reference code, and tests
jointly ground specification $A$.  DeepSeek-V4-Pro receives exactly this one
record and proposes a specification $B$.  The generator is not asked merely to
paraphrase the task or to invent arbitrary failing tests.  It must identify one
named \emph{semantic slot} and change only the value of that slot.  Examples
include strict versus inclusive thresholding, zero- versus one-based indexing,
first- versus last-occurrence retention, case-sensitive versus
case-insensitive matching, and first- versus last-item tie breaking.

The generation contract enforces the following design requirements.
\begin{itemize}
    \item $A$ must preserve the MBPP behavior and $B$ must be the smallest
          plausible counterfactual within the same task family.
    \item Both sides must use the same function name and exact Python
          signature.  If the MBPP name reveals the disputed rule, only the
          name is replaced by a neutral one on both sides.
    \item Input types and domain, output type and structure, ordering and
          duplicate behavior, mutation policy, error behavior, and all
          non-disputed constraints must be shared.
    \item The two documentation strings must be parallel in length, detail,
          specificity, and modal strength.  Neither side may include an extra
          example, edge case, or implementation hint.
    \item The candidate must declare exactly two distinct, valid positional
          input arrays.  On each input, the independently traced outputs must
          satisfy $A(x) \ne B(x)$.
    \item Tasks requiring files, network access, randomness, time, external
          packages, unstable iteration order, uncontrolled floating-point
          behavior, or subjective judgment are excluded.
\end{itemize}
The generator can instead return \texttt{unsuitable} when no honest one-slot
counterfactual exists.  This option is important because forcing every MBPP
record to yield a pair would favor artificial or multi-factor conflicts.

All construction roles used separate calls to
\texttt{deepseek-ai/DeepSeek-V4-Pro} through Together.  Thus, ``separate
evaluator'' below means an independent call and prompt role, not a different
model family.  For the artifact used in the reported experiment, construction
calls used temperature 0.1, seed 42, a maximum of 4,086 completion tokens, a
120-second API timeout, and reasoning disabled.  Network retries did not
consume the three semantic attempts.

\subsection{Semantic validation and repair}

Validation first checks facts that do not require model judgment.  The returned
object must have the exact schema; its copied prompt, source code, tests, and
task identifier must match the selected MBPP record; its signature must parse
as one synchronous function definition with the declared name; and each
discriminating input must be a JSON array with an arity compatible with the
signature.  The two inputs must be distinct, and neither declared case may have
equal $A$ and $B$ outputs.

A second DeepSeek-V4-Pro call then evaluates the candidate on five dimensions,
each scored from 1 to 5: same task family, exactly-one-slot conflict,
information balance, deterministic testability, and clarity.  The evaluator is
instructed to recompute both declared outputs, ask whether a single
implementation could satisfy both documents, check that the function name is
neutral, and reject any hidden second difference.  Acceptance requires a total
score of at least 22, a score of 5 for the single-slot criterion, and scores of
at least 4 on every other dimension.  These thresholds are rechecked by the
host even if the evaluator's textual verdict is inconsistent with its scores.

Rejected candidates follow one of two repair paths.  A \emph{normal repair}
returns the previous candidate and the complete evaluator diagnostics to the
generator, preserving valid content while repairing local balance, clarity,
schema, or output-tracing errors.  A \emph{semantic restart} is used when the
rules are equivalent, the conflict is not observable, or the declared cases
are semantic no-ops.  In that case, the previous $B$, semantic slot, and both
inputs are discarded; only the source grounding and $A$ are retained.  The
generator must propose a genuinely new slot and two new inputs, rather than
editing the claimed outputs.  A source task is abandoned after three malformed
or rejected semantic candidates, or immediately after an
\texttt{unsuitable} response.

\subsection{Executable oracles and pair verification}

An accepted semantic description is not yet an experimental item.  The
pipeline next constructs executable oracles in two stages.  First,
DeepSeek-V4-Pro generates independent reference implementations for $A$ and
$B$.  This prompt requests code only and withholds test-generation duties.  A
reference implementation must use the exact shared signature, implement a
general rule rather than hard-code the declared examples, and reproduce its
side of both discriminating cases.  Host-side AST validation restricts code to
a small allowlist of standard-library functionality and rejects dynamic
execution, reflection, I/O, network or subprocess access, randomness, and
other unsafe or nondeterministic behavior.

Second, after both implementations pass their declared cases, the model
proposes \emph{inputs only}.  Declared discriminating case 0 is automatically
placed in the visible set and case 1 in the hidden set.  The model supplies two
additional visible inputs and five additional hidden inputs, producing three
visible and six hidden calls.  It never supplies expected values or assertion
strings.  The host executes each already validated reference implementation on
the shared inputs and uses the observed return values to form paired assertions
for $A$ and $B$.  Consequently, the expected outputs used for acceptance and
later attribution are execution-derived, not copied from an LLM response.

The host then enforces the following invariants:
\begin{itemize}
    \item all tests have the literal-only form
          \texttt{assert function(literals) == literal};
    \item $A$ and $B$ suites use identical calls in identical order, while
          visible and hidden calls are disjoint and contain no duplicates;
    \item at least one call in each visibility split distinguishes the two
          specifications;
    \item executions occur in fresh isolated \texttt{python -I -S}
          subprocesses and temporary working directories, with a two-second
          timeout; and
    \item every cell is stable across two fresh repetitions.
\end{itemize}
Finally, all eight cells of the solution-by-suite matrix must have the expected
outcome:
\begin{table*}[t]
\centering
\caption{Execution outcomes under the four counterbalanced configurations.}
\label{tab:verification}
\begin{tabular}{lcccc}
\toprule
 & Visible $A$ & Hidden $A$ & Visible $B$ & Hidden $B$ \\
\midrule
Reference solution $A$
& Pass & Pass & Assertion failure & Assertion failure \\
Reference solution $B$
& Assertion failure & Assertion failure & Pass & Pass \\
\bottomrule
\end{tabular}
\end{table*}
Both solutions must also reproduce every declared discriminating output.
Instability, a timeout, a safety violation, a cross-suite pass, or failure on a
same-side suite invalidates the pair.  Oracle diagnostics can trigger targeted
solution or input repair, but the accepted semantic candidate is immutable at
this stage.  Each oracle stage is limited to three rounds.

\subsection{Counterbalanced preference trials and exact target prompt}

For each verified pair, we construct the two genuinely conflicting source
assignments
\[
(\operatorname{doc}(A), \operatorname{tests}(B))
\quad\text{and}\quad
(\operatorname{doc}(B), \operatorname{tests}(A)).
\]
Each assignment is rendered once with documentation first and once with tests
first.  This $2 \times 2$ crossing swaps both the behavioral assignment and the
presentation order, yielding four trials per pair.  No condition labels,
\texttt{spec\_A}/\texttt{spec\_B} identifiers, reference implementations, or
hidden tests are exposed to the target model.

Apart from substituting the documentation, the three visible assertions, the
shared signature, and the order of the two conflicting blocks, the target
prompt is exactly:
\begin{figure*}[h]
\begin{lstlisting}[
  basicstyle=\ttfamily\small,
  breaklines=true,
  breakatwhitespace=true,
  columns=fullflexible,
  keepspaces=true,
  showstringspaces=false,
  frame=single
]
Given the following information:
{documentation and visible tests, in the assigned order}

output the implementation code as a function.

Return your final answer in exactly this format:
<implementation_code>

{shared function signature}

</implementation_code>
\end{lstlisting}
\caption{Prompt template for code-generation experiments.}
\label{fig:code-generation-prompt}
\end{figure*}
There are deliberately no labels such as ``authoritative documentation'' or
``tests take precedence.''  The signature inside the answer scaffold is shared
by both specifications and therefore does not favor either behavior.

\subsection{Target models and reasoning settings}

Each target model receives all 268 trials, with one completion request per
trial.  Table~\ref{tab:mbpp-models} lists the 12 models included in Figure~6.
``No explicit effort'' means that the request omitted a provider-specific
reasoning-effort field and used the endpoint default; it does not mean that the
model performed no internal reasoning.  Conversely, \texttt{high} and
\texttt{max} are API-level inference-budget controls, not a request to reveal a
chain of thought.  Kimi-K3 exposed \texttt{max}, rather than \texttt{high}, as
its supported mandatory-reasoning setting.

\begin{table*}[t]
\centering
\small
\caption{Target models and request settings for the Figure~8 result snapshot.
All requests allowed at most 4,086 completion tokens and used a 120-second API
timeout.  ``Default'' denotes no explicit reasoning-effort parameter.}
\label{tab:mbpp-models}
\begin{tabular}{p{0.20\textwidth}p{0.27\textwidth}p{0.13\textwidth}p{0.21\textwidth}}
\hline
Model & Endpoint model identifier & Reasoning effort & Sampling fields \\
\hline
Claude Opus 4.8 & \texttt{anthropic/claude-opus-4.8} (OpenRouter) & high & temperature and seed omitted \\
DeepSeek-V4-Flash & \texttt{deepseek/deepseek-v4-flash} (OpenRouter) & default & temperature 0.1; seed 42 \\
DeepSeek-V4-Pro & \texttt{deepseek/deepseek-v4-pro} (OpenRouter) & default & temperature 0.1; seed 42 \\
Gemma-3-4B-IT & \texttt{google/gemma-3-4b-it} (OpenRouter) & default & temperature 0.1; seed 42 \\
GPT-4.1-nano & \texttt{gpt-4.1-nano} (OpenAI) & default & temperature 0.1; seed 42 \\
GPT-5-nano & \texttt{gpt-5-nano} (OpenAI) & default & temperature omitted; seed 42 \\
GPT-5.4-nano & \texttt{gpt-5.4-nano} (OpenAI) & high & temperature omitted; seed 42 \\
GPT-5.6-Luna & \texttt{gpt-5.6-luna} (OpenAI) & high & temperature omitted; seed 42 \\
GPT-5.6-Terra & \texttt{gpt-5.6-terra} (OpenAI) & high & temperature omitted; seed 42 \\
Kimi-K3 & \texttt{moonshotai/kimi-k3} (OpenRouter) & max & temperature and seed omitted \\
Llama-3.3-70B-Instruct & \texttt{meta-llama/}\allowbreak\texttt{llama-3.3-}\allowbreak\texttt{70b-instruct} (OpenRouter) & default & temperature 0.1; seed 42 \\
Llama-4-Scout & \texttt{meta-llama/llama-4-scout} (OpenRouter) & default & temperature 0.1; seed 42 \\
\hline
\end{tabular}
\end{table*}

A GPT-OSS-120B result directory is retained with the released experimental
artifacts, but that model is not part of the 12-model Figure~6 analysis and is
not included in the reported correlation with model capability.

\subsection{Implementation extraction, attribution, and aggregation}

The evaluator first extracts the text between the requested implementation
tags (with conservative fallbacks for otherwise valid Python), then checks
that the code defines the required function with the exact signature and
passes the same static safety policy used for the reference implementations.
The extracted implementation is run separately against visible $A$, hidden
$A$, visible $B$, and hidden $B$ suites.  Hidden tests are used for the primary
behavioral attribution:
\begin{itemize}
    \item pass hidden $A$ and fail hidden $B$: follows $A$;
    \item pass hidden $B$ and fail hidden $A$: follows $B$;
    \item pass both hidden suites: compatible with both on the oracle inputs;
    \item fail both hidden suites but pass the displayed visible suite:
          visible-test overfitting; and
    \item fail both hidden suites and the displayed suite, or fail extraction,
          safety, or execution: other/failure.
\end{itemize}
An $A$ or $B$ attribution is finally mapped to \emph{documentation} or
\emph{tests} using that trial's source assignment.  Thus the label does not
assume that $A$ is always the documentation-supported behavior.  The
test-preference rate in Figure~6 is
\[
\frac{N_{\mathrm{tests}}}
     {N_{\mathrm{tests}} + N_{\mathrm{documentation}}},
\]
so both-compatible, visible-test-overfit, wrong-behavior, extraction, and
execution outcomes are excluded from this conditional preference denominator
but remain available in the released raw result records.  Confidence intervals
are obtained from 10,000 bootstrap resamples clustered by conflict-pair
identifier, keeping all four counterbalanced trials from a sampled pair
together. The results are shown in Table \ref{tab:mbpp-outcome-summary}.

\begin{table*}[t]
\centering
\small
\caption{Code-generation outcomes by target model. Documentation and Tests
are conditional percentages among trials attributed to exactly one of the two
conflicting specifications. Overfit and Failure are percentages of all 268
trials per model. Overfit denotes an implementation that passes the displayed
visible-test suite but fails both hidden suites; Failure denotes an extraction,
safety, runtime, or API hard failure. GPT-OSS-120B is excluded to match the
12-model analysis.}
\label{tab:mbpp-outcome-summary}
\begin{tabular}{lrrrr}
\hline
Model & Documentation & Tests & Overfit & Failure \\
\hline
Claude Opus 4.8          &  2.9\% & 97.1\% & 5.6\% &  2.6\% \\
DeepSeek-V4-Flash        & 42.4\% & 57.6\% & 3.0\% &  5.2\% \\
DeepSeek-V4-Pro          & 21.6\% & 78.4\% & 1.5\% &  8.6\% \\
Gemma-3-4B-IT            & 83.5\% & 16.5\% & 2.6\% &  1.9\% \\
GPT-4.1-nano             & 82.6\% & 17.4\% & 1.5\% &  5.6\% \\
GPT-5-nano               & 14.3\% & 85.7\% & 4.5\% &  8.6\% \\
GPT-5.4-nano             & 32.1\% & 67.9\% & 6.0\% &  3.7\% \\
GPT-5.6-Luna             & 25.7\% & 74.3\% & 5.2\% &  1.9\% \\
GPT-5.6-Terra            & 30.7\% & 69.3\% & 3.0\% &  1.1\% \\
Kimi-K3                  &  6.0\% & 94.0\% & 2.6\% & 15.7\% \\
Llama-3.3-70B-Instruct   & 65.3\% & 34.7\% & 1.5\% &  4.5\% \\
Llama-4-Scout            & 71.8\% & 28.2\% & 1.1\% & 18.3\% \\
\hline
\end{tabular}
\end{table*}

\section{Clinical Rule Conflict}
\label{app:medical}

This appendix describes the construction of the clinical rule conflict experiment summarized in
Section \emph{Clinical Rule Conflicts}, and reports the cross-authority analysis and robustness checks that
are not carried in the main text.

\subsection{Dataset construction}

The conflict material is derived from public-domain drug labelling retrieved through the openFDA drug
label API. For a
candidate label, a language model \emph{proposes} a dosing or indication threshold together with a
verbatim supporting quote; the proposal carries no authority of its own and is accepted only if it
passes a deterministic gate. The gate requires that the quoted span contain the metric, the numeric
value, and an explicit below-relation, and rejects dose-adjustment bands, wrong-direction
statements, fragments of dosing tables, and ``not studied'' prose. Its rejection rules were
extended over three rounds in which sampled acceptances were re-examined against the source text and
each observed failure mode was turned into an additional rule. Roughly $17\%$ of candidate labels
survive the gate. Accepted thresholds cover two clinical decision fields, renal dosing thresholds
and pediatric minimum-age indications, yielding 166 conflict instances. Every accepted instance
retains its quote and source identifier, so any threshold can be checked against the label it came
from.

Each instance pairs the genuine threshold with a schema-preserving perturbation of its value,
following the procedure used to construct $f_2$ in the mathematical benchmark, and a query is placed
strictly between the two values so that the rules give opposite answers. The perturbation is applied
in \emph{both} directions: the genuine label value is the more restrictive of the two competing rules
in roughly half of the instances and the more permissive one in the other half. This is what allows a
preference for restrictiveness to be distinguished from a preference for correctness in
Section \emph{Clinical Rule Conflicts}; were the perturbation always to loosen the genuine threshold,
``adopts the stricter rule'' and ``adopts the genuine rule'' would denote the same event.

Each of the two rules is then rendered in all four representation types of the main study.
Table~\ref{tab:medical-renderings} shows a complete instance. The genuine label value is $12$ years
and the perturbation is $17$ years, so here the \emph{perturbed} rule is the more restrictive one;
the query at age $14$ separates them.

\begin{table*}[t]
\centering
\small
\caption{One complete conflict instance (acetazolamide extended-release, pediatric minimum age).
Query: \emph{``For a 14-year-old patient, is use of the medicine indicated? Answer yes or no.''} The
left column is the genuine labelled threshold, the right column its perturbation; each row is one
representation type. A trial presents one cell from each column.}
\label{tab:medical-renderings}
\begin{tabular}{@{}p{0.11\textwidth}p{0.42\textwidth}p{0.42\textwidth}@{}}
\toprule
& \textbf{Genuine rule} (threshold 12; answers \emph{yes})
& \textbf{Perturbed rule} (threshold 17; answers \emph{no}) \\
\midrule
Formal
& $\text{use}(a) = \begin{cases} \text{indicated} & a \ge 12 \\ \text{not established} & a < 12 \end{cases}$
& $\text{use}(a) = \begin{cases} \text{indicated} & a \ge 17 \\ \text{not established} & a < 17 \end{cases}$ \\[2.2em]
Naturalized formal
& Use is indicated for patients aged $\ge 12$ years, and not established for patients aged $< 12$ years.
& Use is indicated for patients aged $\ge 17$ years, and not established for patients aged $< 17$ years. \\[0.6em]
Pure natural language
& The medicine is indicated for patients who are at least twelve years old; for children younger than that, its use is not established.
& The medicine is indicated for patients who are at least seventeen years old; for children younger than that, its use is not established. \\[0.6em]
Input--output examples
& age 16 years $\rightarrow$ indicated; age 8 years $\rightarrow$ not established; age 22 years $\rightarrow$ indicated.
& age 21 years $\rightarrow$ indicated; age 13 years $\rightarrow$ not established; age 27 years $\rightarrow$ indicated. \\
\bottomrule
\end{tabular}
\end{table*}

The example sets exclude the queried value, so the examples never state the answer outright and the
model must recover the underlying threshold from the demonstrations. All four renderings of a rule
express identical semantics; only the surface form differs. This is also the most plausible reason
for the weakness of examples reported in Section \emph{Clinical Rule Conflicts}: clinical rules involve
compound conditions, multiple thresholds, and exceptions, which a finite set of demonstrations
covers poorly.

\subsection{Trials, models, and attribution}

Each conflict is evaluated on all six unordered representation pairs, under both authority framings
(``Source 1/2'' versus ``Clinical guideline''/``Product label''), and under a complete $2\times2$
symmetry: presentation order is reversed, and the assignment of representation type to underlying
rule is swapped. Order and rule-identity effects therefore cancel in the aggregate. The full design
is $166 \times 6 \times 2 \times 2 \times 2 = 7{,}968$ trials per model.

The nine evaluated models span five families: Doubao Seed~2.0 Lite and Pro, GPT-4.1-nano,
GPT-5-nano, GPT-5.4-nano, DeepSeek-V4-Flash and V4-Pro, Llama-3.3-70B, and Llama-4-Scout. We report
only effects that hold in the same direction for all nine.

Running the complete design on every model is prohibitive for verbose reasoners, so coverage is
uneven and is stated wherever results are reported. Doubao Seed~2.0 Lite was run on the complete
design; six models were run on a fixed $1{,}500$-trial subsample (seed 42); DeepSeek-V4-Pro on a
$700$-trial subset nested inside that subsample; and Doubao Seed~2.0 Pro on an earlier $2{,}147$-trial
sample that overlaps the common subsample only partially (413 trials), so it is reported alongside
rather than as part of the matched comparison. Every model has the full $372$ cross-authority trials.
All models were reached through a single endpoint with identical request settings
(\texttt{reasoning = medium} where accepted, temperature $0.1$); no API errors occurred in any run.

Because the query is positively phrased and the two rules answer it oppositely by construction, a
clean \emph{yes} or \emph{no} maps deterministically to one rule. Attribution is recomputed from the
stored responses at analysis time by a parser that tolerates verdicts wrapped in \LaTeX{}
(\verb|\text{no}|, \verb|\boxed{Yes}|) or preceded by prose. Responses the parser cannot resolve are
resolved by a separate adjudication pass that classifies the stated position as following one rule,
the other, both, or neither. The adjudication uses a model from a family not among those evaluated,
so that no model grades itself, and every label is released with the response it was assigned to, so
the classification can be re-derived or replaced.

\subsection{Cross-authority conflicts}
\label{app:medical-authority}

The constructed conflicts use genuine values but an anonymised generic rule, because attaching a real
agency's name to a perturbed rule would fabricate a regulatory claim. To examine conflicts between
named authorities we assembled a separate set of 31 instances in which the U.S.\ FDA and the European
EMA issue operationally opposite guidance for the same drug and patient. Each instance carries, for
both regulators, a one-line rule, a verbatim quote, and a resolvable citation---DailyMed or
accessdata.fda.gov for the FDA, and EPAR product information or SmPC from the EMA medicines
database. Candidates were
collected and then subjected to a refutation check: for each candidate, the claimed conflict was
challenged against the cited primary documents, and a candidate was retained only if the challenge
failed. Two candidates did not survive and were dropped. The check screens candidates for exclusion
rather than warranting the survivors---the quotes and citations it operated on are released with the
dataset, so every conflict can be verified directly against the labelling. Because a rule cannot be
truthfully re-attributed to the other authority, assignment symmetry is replaced by direction
balance: 17 instances are FDA-more-permissive and 14 EMA-more-permissive. Each conflict is presented
under a labeled and an anonymized condition, with order counterbalanced and three repetitions, giving
$372$ trials per model.

\begin{figure}[t]
    \centering
    \includegraphics[width=0.92\linewidth]{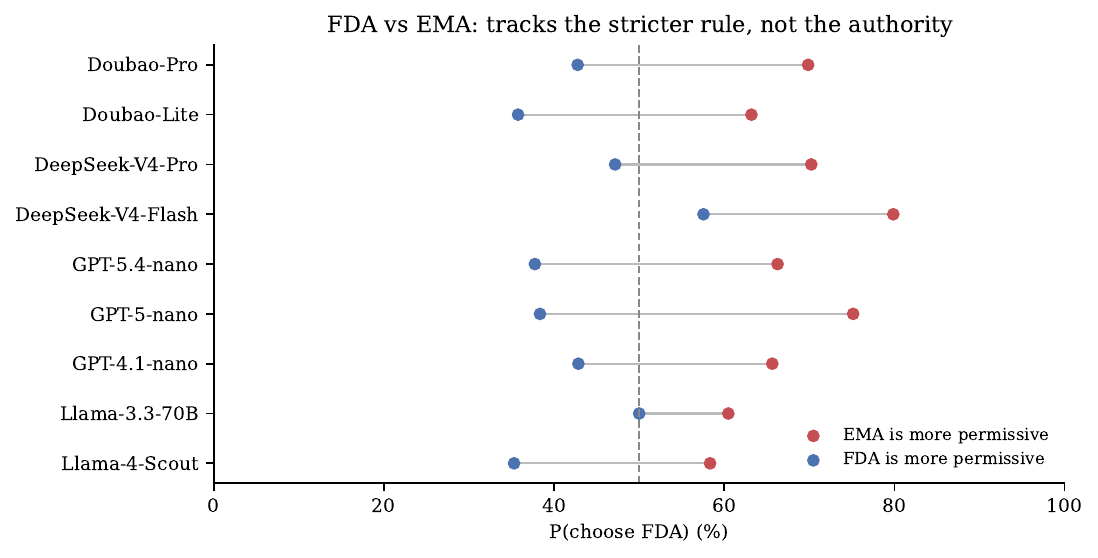}
    \caption{Cross-authority conflicts. Probability of choosing the FDA rule, split by which
    authority is the more permissive one. The gap is positive for all nine models, so the
    near-balanced aggregate reflects two opposing biases rather than indifference.}
    \label{fig:medical-authority}
\end{figure}

Direction balance permits the analysis in Figure~\ref{fig:medical-authority}. In aggregate no model
prefers one regulator (FDA-selection $46\%$--$68\%$), but for every model this is the average of two
opposing biases: the probability of choosing the FDA rule is lower when the FDA rule is the more
permissive one and higher when the EMA rule is. The direction-split swing is positive for all nine
models, ranging from $+10$ to $+37$ percentage points. Restrictive selection is $55\%$--$68\%$. The
restrictiveness preference reported in Section \emph{Clinical Rule Conflicts} therefore also holds on real,
dual-cited regulatory divergences containing no fabricated values.

Naming the authorities has a markedly model-specific effect. Hedging---reporting both positions
without committing to either---rises sharply for some families when the regulators are named and not
at all for others (Table~\ref{tab:medical-hedging}). The Doubao and DeepSeek families hedge
considerably more once the sources are named, whereas the GPT and Llama families are essentially
unaffected and commit under either framing. We therefore report this as a property of particular
models rather than as a general effect of authority framing.

\begin{table}[t]
\centering
\small
\caption{Hedging rate (reports both positions, commits to neither) by condition. The effect of
naming the authorities is confined to particular model families.}
\label{tab:medical-hedging}
\begin{tabular}{@{}lrr@{}}
\toprule
Model & Anonymized & Labeled \\
\midrule
Doubao Seed 2.0 Lite & 11\% & 30\% \\
Doubao Seed 2.0 Pro  & 4\%  & 32\% \\
DeepSeek-V4-Flash    & 4\%  & 24\% \\
DeepSeek-V4-Pro      & 19\% & 25\% \\
GPT-4.1-nano         & 0\%  & 3\%  \\
GPT-5-nano           & 1\%  & 6\%  \\
GPT-5.4-nano         & 0\%  & 1\%  \\
Llama-3.3-70B        & 1\%  & 1\%  \\
Llama-4-Scout        & 0\%  & 0\%  \\
\bottomrule
\end{tabular}
\end{table}

This analysis measures preference between \emph{sources}, not between representations. It is a
complement to the main line of the paper rather than evidence for it, which is why it is reported
here.

\subsection{Robustness}

\paragraph{Dataset screening.} Before any result was computed, all 31 cross-authority instances were
screened against five explicit criteria: whether each quote supports the rule attributed to it,
whether the query is genuinely discriminative given the two rules, whether the stated answer polarity
follows from the rule, whether the recorded direction matches the answers, and whether the claim is
consistent with known labelling. Two of these criteria---answer polarity and direction---are
mechanical, and were re-derived deterministically in code; both were consistent for all 31 instances.
Of the remaining checks, 28 instances passed cleanly and 3 were flagged, none of them as a
fabrication. In each flagged case the two sides are not a perfect like-for-like pairing: an
indication-specific sub-label set against an indication-general table, a route mismatch between an
oral and a parenteral product, and one regulatory action that could not be corroborated from the
cited material alone. All three are retained, and a sensitivity check recomputes the headline
contrasts without them. Screening identifies candidates for exclusion; it does not itself warrant
correctness. The quote and citation underlying every rule are released with the dataset, so each
instance can be checked against the primary document independently of this procedure.

\paragraph{Prompt contamination.} Two defects were found in the mined cross-authority instances and
fixed before any number was reported. First, three queries had the answer appended by the mining
process (``FDA answer: NO. EMA answer: YES''), which leaks the label into the prompt, including in
the anonymized condition. Second, seven rules and two queries identified their own regulator through
incidental wording (``SmPC section 4.3'', ``FDA-approved test''), which silently de-anonymizes the
source in precisely the condition designed to hide it. The encoder now strips answer statements and
emits a scrubbed parallel rendering used only in the anonymized condition, asserting that no
regulator token survives. The pre-fix run was discarded and every reported result comes from the
post-fix re-run.

\paragraph{Heterogeneity in the magnitude of the restrictiveness preference.}
The direction of the restrictiveness preference reported in
Section \emph{Clinical Rule Conflicts} is uniform across all nine models, but its
magnitude varies substantially across instances: the two clinical fields differ
by 15 to 35 percentage points, and this difference has the same sign for every
model. We report this heterogeneity as an observation and do not attribute it to
a cause. The two fields differ simultaneously in rule type, in the wording of
the restrictive outcome (\emph{contraindicated} versus \emph{not established}),
and in the magnitude of the perturbation, and a design with two fields cannot
separate these factors. Identifying what modulates the strength of the
preference requires a dedicated design and is left to future work.

\paragraph{Response truncation.} Verbose reasoning models initially exhausted their token budget
before emitting a verdict on about $10\%$ of trials, and the loss was not uniform across clinical
fields. The budget was raised and the affected trials rerun; residual unresolved trials are now
$0$--$2\%$ per model and are excluded from rates rather than imputed.

\section{Additional Results}
\label{app:additional-results}

\subsection{Example Selection across Function Families}

\begin{figure*}[t]
    \centering
    \includegraphics[width=\textwidth]{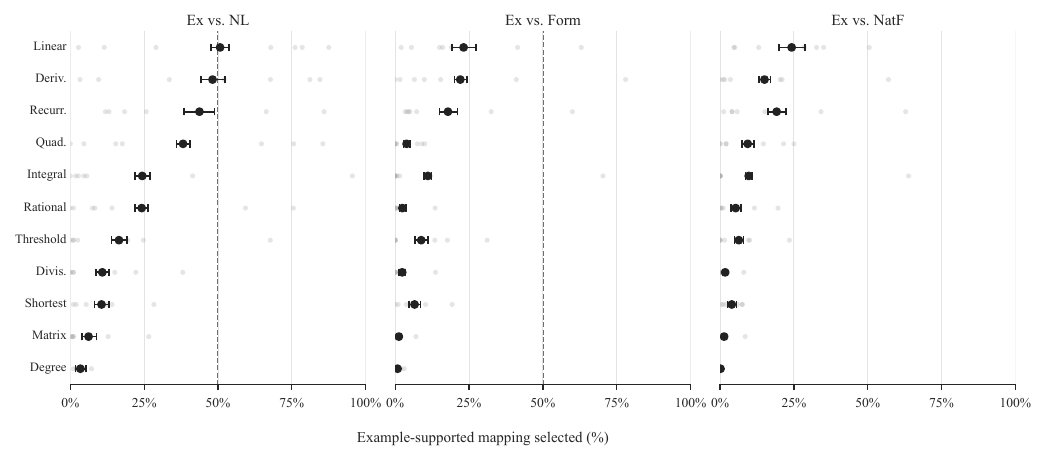}
\caption{Example selection by function family across all three
example-based comparisons. Light points show model-specific estimates,
and dark points show pooled estimates across models. Each estimate is
the proportion of attributable responses selecting the
example-supported mapping. Error bars show conflict-instance
clustered-bootstrap \(95\%\) confidence intervals.}
    \label{fig:example-use-by-family-appendix}
\end{figure*}

Figure~\ref{fig:example-use-by-family-appendix} shows that
example use varies substantially across mathematical function
families. Examples are most competitive against pure natural-language
specifications for linear functions, derivative queries, and recurrence
sequences, where several pooled selection rates approach or exceed
\(50\%\). Their influence is much weaker for graph-degree queries,
matrix computation, shortest-path problems, and several conditional
and rational-function families.

The same family-level ordering is broadly preserved when examples
compete with formal or naturalized-formal specifications, but the
selection rates shift downward across nearly all families. This shows
that example use is determined jointly by the underlying task structure
and the representation of the competing specification. Examples become
competitive when finite demonstrations expose a recoverable mapping,
but they are consistently disadvantaged when the alternative presents
the rule through explicit mathematical structure.

\subsection{Presentation-Order Effects}

\begin{figure}[t]
    \centering
    \includegraphics[width=\columnwidth]
        {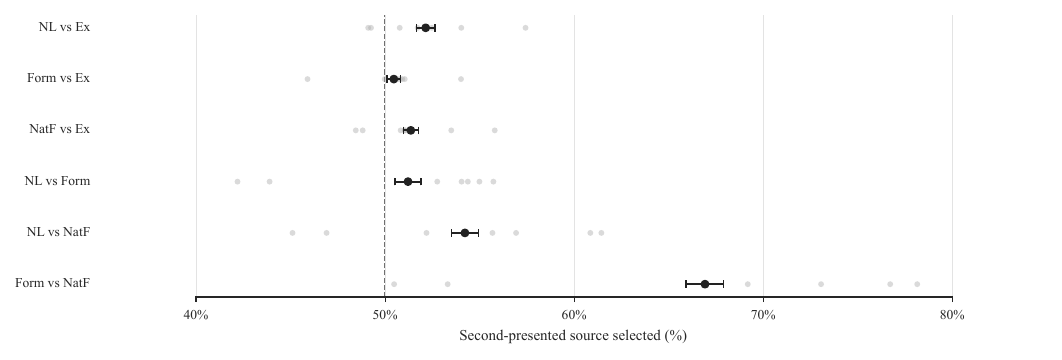}
    \caption{Presentation-order effects across representation comparisons.
    Light gray points show model-specific estimates, and black points show
    pooled estimates across models. Each estimate is the proportion of
    attributable responses selecting the specification presented second.
    Horizontal error bars show conflict-instance clustered-bootstrap
    \(95\%\) confidence intervals for the pooled estimates, and the dashed
    line marks \(50\%\).}
    \label{fig:presentation-order-effects}
\end{figure}

Figure~\ref{fig:presentation-order-effects} shows that presentation
order has a measurable but generally secondary effect. For five of the
six representation comparisons, the pooled probability of selecting
the second-presented specification remains relatively close to
\(50\%\), and these order effects are small compared with the
corresponding representation preferences.

The formal versus naturalized-formal comparison is the clear
exception. These two representations have similar overall selection
strength, with neither exerting the large representation advantage
observed in comparisons involving pure natural language or examples.
Consequently, when the representation-level preference is weak,
secondary decision factors become more influential, and the
second-position preference is amplified. This result suggests that
presentation order plays the largest role when the competing
specifications are otherwise similarly preferred.

\subsection{Non-attributable Responses}

\begin{figure*}[t]
    \centering
    \includegraphics[width=\textwidth]{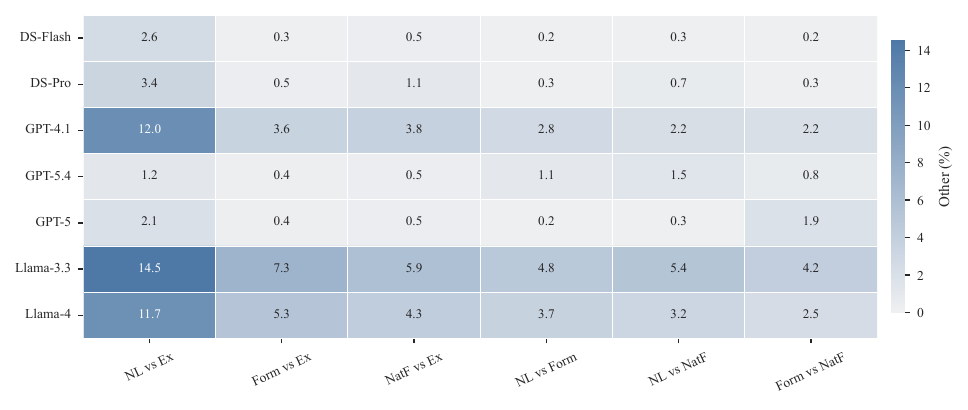}
\caption{Rates of responses assigned to \emph{other} across models and
representation comparisons. Each cell reports the percentage of all
trials whose normalized final answer matches neither executable
candidate output. Non-attributable responses are generally rare and
are concentrated in the pure-natural-language versus examples
condition.}
    \label{fig:other-response-rates}
\end{figure*}

Figure~\ref{fig:other-response-rates} shows that responses assigned to
\emph{other} are uncommon in most representation comparisons. For
formal, naturalized-formal, and their comparisons with examples, the
rate is generally below \(5\%\), indicating that model outputs can
usually be attributed cleanly to one of the two executable mappings.

The main exception is the pure-natural-language versus examples
condition, particularly for GPT-4.1-nano and the Llama-family models.
Manual inspection of these responses shows that they do not combine
the two competing mappings. Instead, the model typically fails to infer
the underlying mapping from the finite input--output examples or
applies the inferred rule incorrectly to the discriminative query.
The elevated \emph{other} rate therefore reflects limited example-based
rule induction and execution capability, rather than ambiguity in the
attribution procedure or the emergence of hybrid behaviors.

\subsection{Boolean Algebra}
\begin{figure}[t]
    \centering
    \includegraphics[width=0.8\linewidth]{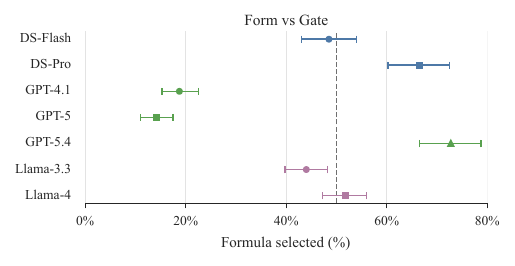}
    \caption{
Results on the Boolean Algebra dataset, showing model preferences between
Boolean formulas and gate netlists. Points indicate the proportion of
attributable responses selecting the formula-supported mapping, with
clustered-bootstrap \(95\%\) confidence intervals over conflict instances.
}
    \label{fig:bool_formula_vs_gate}
\end{figure}

\begin{figure}[t]
    \centering
    \includegraphics[width=0.8\linewidth]{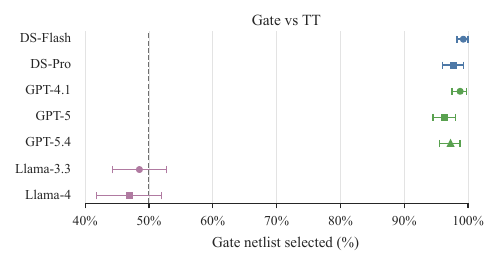}
\caption{
Results on the Boolean Algebra dataset, showing model preferences between
gate netlists and truth tables. Points indicate the proportion of
attributable responses selecting the gate-netlist-supported mapping, with
clustered-bootstrap \(95\%\) confidence intervals over conflict instances.
}    \label{fig:bool_gate_vs_table}
\end{figure}

Figure \ref{fig:bool_formula_vs_gate} compares Boolean formulas with gate netlists. Unlike the strong aggregate preference observed between formulas and truth tables, this comparison does not reveal a consistent ordering across models. GPT-4.1 and GPT-5 strongly favor the gate-netlist-supported mapping, selecting the formula-supported mapping in only approximately 19\% and 14\% of attributable responses, respectively. In contrast, DeepSeek-V4-Pro and GPT-5.4 favor formulas, with formula selection rates of approximately 67\% and 73\%. DeepSeek-V4-Flash, Llama-3.3, and Llama-4 remain closer to the 50\% indifference point, although Llama-3.3 exhibits a modest preference for gate netlists. These substantial cross-model differences suggest that formulas and gate netlists occupy a more comparable level of representational strength: both encode the Boolean function compositionally and support direct evaluation, leaving the preferred representation more dependent on model-specific training or processing biases.

Figure \ref{fig:bool_gate_vs_table} shows a markedly different pattern when gate netlists are compared with truth tables. All DeepSeek and GPT models select the gate-netlist-supported mapping in more than 95\% of attributable responses, with confidence intervals well above the 50\% indifference point. This indicates a highly consistent preference for the gate-netlist representation over the truth table among these models. The two Llama models are exceptions, producing estimates close to chance with confidence intervals overlapping 50\%, and therefore show no reliable preference between the two representations.

Taken together, the two comparisons refine the broader ordering among Boolean representations. Gate netlists consistently dominate truth tables for most evaluated models, but their relationship with formulas is model-dependent rather than universal. One plausible interpretation is that both formulas and gate netlists provide compact, intensional descriptions of the underlying Boolean function, whereas truth tables present the function extensionally as a collection of input-output assignments. The observed preference for gate netlists over truth tables may therefore reflect sensitivity to compositional structure, compactness, or ease of query-specific evaluation, rather than a general preference for any particular surface syntax. However, because these properties are not independently controlled in the present experiment, this interpretation should be treated as suggestive rather than causal.

These results further demonstrate that representational preference is not fully captured by a single global hierarchy: robust ordering emerges between some representation pairs, while semantically equivalent structural representations can produce pronounced model-specific reversals.

\section{Abstention-Enabled Conflict Resolution}
\label{app:abstention}

To examine whether models always resolve conflicting specifications by
committing to one of the competing sources, we conduct an additional
abstention-enabled experiment. We use the same conflict instances,
representation pairs, counterbalanced configurations, and evaluation setup as
in the main mathematical experiments.

The only modification is to the final instruction in the prompt. In the main
experiment, the prompt ends with:

\begin{quote}
\texttt{Compute the value: [query]. Please give your final answer in the form
\textbackslash boxed\{...\}.}
\end{quote}

For the abstention-enabled experiment, we minimally modify this instruction to:

\begin{quote}
\texttt{Compute the value: [query]. Please give your final answer in the form
\textbackslash boxed\{...\}, or answer "uncertain".}
\end{quote}

Thus, the model receives exactly the same conflicting specifications and query,
but is additionally given an explicit option to abstain from selecting either
candidate behavior.

We evaluate DeepSeek-V4-Pro and GPT-5-nano under this setting. As in the main
experiment, each model is evaluated on all six representation pairs and four
counterbalanced configurations for every conflict instance, resulting in
13,200 trials per model. We define the abstention rate as the proportion of
all trials in which the model returns the explicit \emph{uncertain} response:
\[
P(\mathrm{abstain})
=
\frac{N_{\mathrm{uncertain}}}{N_{\mathrm{all\ trials}}}.
\]

Table~\ref{tab:abstention} reports abstention rates for each representation
pair. DeepSeek-V4-Pro abstains in approximately 6\% of trials overall, with
relatively moderate variation across conditions. GPT-5-nano exhibits a higher
overall abstention rate of approximately 17\%, but this behavior varies
substantially across competing source types. In particular, GPT-5-nano
abstains frequently when both sources are textual or formalized
representations, whereas abstention remains uncommon in conditions involving
input--output examples.

\begin{table}[!t]
\centering
\small
\begin{tabular}{lcc}
\toprule
Conflict Pair & DeepSeek-V4-Pro & GPT-5-nano \\
\midrule
Pure NL vs.\ Formal       & 1.59\%  & 15.06\% \\
Pure NL vs.\ Examples     & 6.91\%  & 1.59\%  \\
Formal vs.\ Examples      & 9.41\%  & 1.05\%  \\
NatF vs.\ Examples        & 4.64\%  & 0.82\%  \\
Pure NL vs.\ NatF         & 3.14\%  & 21.92\% \\
Formal vs.\ NatF          & 10.00\% & 64.18\% \\
\midrule
Overall                    & 5.95\%  & 17.45\% \\
\bottomrule
\end{tabular}
\caption{
Abstention rates when models are explicitly allowed to answer
\emph{uncertain}. Rates are computed over all trials in each representation
pair.
}
\label{tab:abstention}
\end{table}

These results indicate that explicit abstention can constitute an additional
strategy for handling conflicting specifications rather than committing to
either competing source. Importantly, abstention does not appear to behave as
a uniform model-level tendency. For GPT-5-nano, it varies sharply with the
types of specifications in conflict, whereas DeepSeek-V4-Pro exhibits a more
moderate pattern. This suggests that abstention may capture a distinct
dimension of conflict-handling behavior that is not reducible to source
selection alone.

\end{document}